\documentclass[times,twocolumn,final]{elsarticle}

\usepackage{medima_reduced}
\usepackage{framed,multirow}

\usepackage{amssymb}
\usepackage{amsmath}
\usepackage{mathtools}
\usepackage{latexsym}

\DeclareMathOperator*{\argmin}{arg\,min}
\DeclareMathOperator{\diag}{diag}
\DeclareMathOperator{\tr}{tr}
\DeclareMathOperator{\cov}{cov}
\DeclareMathOperator{\var}{var}
\DeclareMathOperator{\vecz}{vec}
\DeclareMathOperator{\SO}{SO}
\DeclareMathOperator*{\argmax}{arg\,max}

\usepackage{stfloats}
\usepackage{fancyhdr}

\newcommand{\changefont}{%
    \fontsize{9}{11}\selectfont
}

\usepackage{url}
\usepackage{xcolor}

\usepackage{hyperref}

\definecolor{newcolor}{rgb}{.8,.349,.1}

\begin{document}

\begin{frontmatter}
\title{\textbf{Learning the shape of female breasts: an open-access 3D statistical shape model of the female breast built from 110 breast scans}}%

\author[1,2]{Maximilian \snm{Weiherer}}
\author[3,4]{Andreas \snm{Eigenberger}}
\author[2]{Bernhard \snm{Egger}}
\author[4]{Vanessa \snm{Br\'{e}bant}}
\author[4]{Lukas \snm{Prantl}}
\author[1,5]{Christoph \snm{Palm}\corref{cor1}}
\cortext[cor1]{Corresponding author. \\ \hspace*{4.4mm} \textit{E-mail address}: christoph.palm@oth-regensburg.de (C. Palm).}

\address[1]{Regensburg Medical Image Computing (ReMIC), Ostbayerische Technische Hochschule Regensburg (OTH Regensburg), Regensburg, Germany}
\address[2]{Chair of Visual Computing, Friedrich-Alexander-Universität Erlangen-Nürnberg (FAU), Erlangen, Germany}
\address[3]{Faculty of Mechanical Engineering, OTH Regensburg, Regensburg, Germany}
\address[4]{University Center of Plastic, Aesthetic, Hand and Reconstructive Surgery, University Hospital Regensburg, Regensburg, Germany}
\address[5]{Regensburg Center of Biomedical Engineering (RCBE), OTH Regensburg and Regensburg University, Regensburg, Germany}

\begin{abstract}
We present the \textit{Regensburg Breast Shape Model} (RBSM) -- a 3D statistical shape model of the female breast built from 110 breast scans acquired in a standing position, and the first publicly available.
Together with the model, a fully automated, pairwise surface registration pipeline used to establish dense correspondence among 3D breast scans is introduced.
Our method is computationally efficient and requires only four landmarks to guide the registration process.
A major challenge when modeling female breasts from surface-only 3D breast scans is the non-separability of breast and thorax.
In order to weaken the strong coupling between breast and surrounding areas, we propose to minimize the \textit{variance} outside the breast region as much as possible.
To achieve this goal, a novel concept called \textit{breast probability masks} (BPMs) is introduced.
A BPM assigns probabilities to each point of a 3D breast scan, telling how \textit{likely} it is that a particular point belongs to the breast area.
During registration, we use BPMs to align the template to the target as accurately as possible \textit{inside} the breast region and only roughly outside.
This simple yet effective strategy significantly reduces the unwanted variance outside the breast region, leading to better statistical shape models in which breast shapes are quite well decoupled from the thorax.
The RBSM is thus able to produce a variety of different breast shapes as independently as possible from the shape of the thorax.
Our systematic experimental evaluation reveals a generalization ability of 0.17 mm and a specificity of 2.8 mm.
To underline the expressiveness of the proposed model, we finally demonstrate in two showcase applications how the RBSM can be used for surgical outcome simulation and the prediction of a missing breast from the remaining one. Our model is available at \url{https://www.rbsm.re-mic.de/}.
\end{abstract}

\end{frontmatter}


\section{Introduction}
Since the seminal work of Cootes et al. \cite{Coo95}, \textit{statistical shape models} became an emerging tool to capture natural shape variability within a given class of objects.
As a result, a large number of shape models were developed during the last decades.
The probably most well-known models were built for the human face using textured 3D face scans.  
Introduced by Blanz and Vetter \cite{Bla99}, this class of statistical shape models is commonly known as \textit{3D Morphable Models} (3DMMs) and includes models such as the \textit{Basel Face Model} (BFM) and \textit{Large Scale Facial Model} (LSFM) presented by Paysan et al. \cite{Pay09} and Booth et al. \cite{Boo16}, respectively.
Well-studied applications of 3DMMs include face recognition, expression transfer between individuals, face animation, and 3D face reconstruction from a single 2D photograph \cite{Egg20}.
Especially in the last decade, shape analysis also gained popularity in the field of computational anatomy, where statistical shape models are successfully used to model variations of anatomical objects such as bones and organs. 
Later, these models are utilized for a variety of medical applications including (but not limited to) image segmentation, surgical simulation, therapy planning, and motion analysis \cite{Amb19}.
Despite the popularity of statistical shape models in the aforementioned areas, to date and to the best of our knowledge, no publicly available 3D statistical shape model of the female breast exists.

With breast cancer being the most common malignant neoplasm among women \cite{Sun21}, successful \textit{breast reconstruction surgery} (BRS) is crucial for patients undergoing mastectomy.
In order to give patients a first impression about what their breast might look like after BRS, surgical outcomes are more and more often simulated using patient-specific 3D breast scans, acquired in a standing position (see Section \ref{sec:related_work} for an overview).
Typically, simulations are performed using physically motivated deformable models of the breast.
While these models take into account material properties and physical effects such as gravity, they may \textit{not always} produce realistic-looking shapes as no prior knowledge in form of example shapes is included \cite{Roo06,Su11}.
Hence, simulated outcomes might be physically plausible, but definitely lack \textit{statistical plausibility} in the sense that generated breast shapes are somehow \textit{likely} or similar to those typically observed within a target population.
As the ultimate goal of BRS is an outcome looking as \textit{natural} as possible, we believe that simulation of surgical outcomes must not only rely on physically based deformable models but also should take into account statistical effects.
Indeed, the phenomenon that humans tend to compare themselves with others clearly underlines the importance of the fact that simulated breast shapes should look similar to the breasts within the target population.
As a first step towards combining physical \textit{and} statistical plausibility of simulated breast surgery outcomes, we propose to use 3D statistical shape models built from natural-looking female breasts.
In addition, by introducing such models into the breast shape domain, many of the aforementioned applications from other domains could be transferred to the breast as we will exemplary show later in this article.

To this end, this paper introduces the first publicly available 3D statistical shape model of the female breast built from 110 breast scans acquired in a standing position.
Together with the model, we present a fully automated, pairwise registration pipeline especially tailored for 3D breast scans and its application in the context of statistical shape modeling.
Our method is computationally efficient and requires only four landmarks to guide the registration process.

\subsection{Challenges}
Compared to shape modeling of most parts of the human body, building a statistical shape model for the female breast imposes some new challenges as discussed in the following.

\begin{figure}[b]
\includegraphics{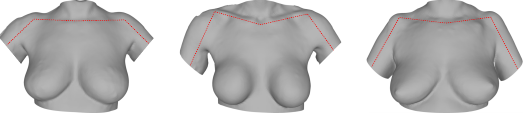}
\caption{Three typical 3D breast scans sampled from our database.
Although a common pose was declared during data acquisition, a lot of pose variations are still present (indicated through a skeleton drawn in red). 
These mainly emerge from the arms and shoulders.}
\label{fig:database_samples}
\end{figure}

\textbf{Data acquisition.}
Firstly, acquiring a sufficient amount of high-quality training data is challenging.
Usually, 3D scanning and \textit{manual} landmark detection is an uncomfortable situation for the participants, in which their upper body is required to be naked.
Moreover, landmarks can be identified only through palpation and by using a regular tape measure, both requiring a physical examination in a clinical environment.
In addition, during the \textit{whole} examination, a specified posture needs to be held fixed ensuring a similar pose across all subjects.
This can be very exhausting, especially if 3D breast scans are taken in a standing position in which both arms should ideally be held away from the body in order to capture the breast as isolated as possible.
As a result, 3D scanning protocols used in clinical practice are often designed to be carried out relatively fast, thus lacking necessary precision for pose standardization, see Figure \ref{fig:database_samples}.
Note that the problem of only \textit{quasi-similar} postures was also recently observed by Mazier et al. \cite{Maz21}.
All in all, the aforementioned factors definitely hinder the implementation of large-scale, high-quality data surveys and might also explain why no publicly available data set of female 3D breast scans exits.

\begin{figure}[t]
\includegraphics[width=88.5mm]{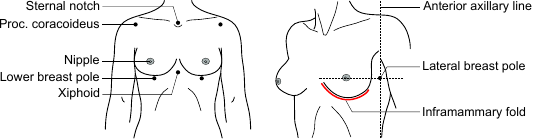}
\caption{A brief overview of common landmarks and key anatomical structures of the female breast and thorax (illustration adapted from \cite{Hwa15}).}
\label{fig:anatomy}
\end{figure}

\textbf{Correspondence estimation.}
Secondly, establishing dense correspondence among 3D breast scans by means of surface registration (rigid and non-rigid) is difficult due to the lack of reliable landmarks.
In essence, only four valid landmarks can be used for non-rigid registration.
These include both nipples as well as both lower breast poles (throughout this work defined as \textit{the lowest (most caudal) point of the breast}, see Figure \ref{fig:anatomy}). 
Anatomical landmark points such as the sternal notch or processus coracoideus cannot be used for non-rigid registration purposes as they would recover undesired, pose-dependent shape variations during statistical shape modeling.
This way, the processus coracoideus disqualifies as its position depends strongly on the position of the arms and shoulders.
On the other hand, the xiphoid, located at the center of the thorax, \textit{could} technically be used for registration. 
However, it cannot be reliably and consistently located across a wide range of differently-shaped breasts because of its dorsal tilt, effectively hindering the identification of a unique point.
Another fundamental problem is the complete lack of robust landmarks on the lateral part of the breast. 
Although Hartmann et al. \cite{Har20} defined a lateral breast pole as the orthogonal intersection between the anterior axillary line and a line passing through the nipple, this point cannot be used for non-rigid registration as its position is also affected by the pose of the arms and shoulders.
Finally, already the initial, rigid alignment of 3D breast scans is challenging due to the lack of reliable landmarks that do not undergo large soft tissue deformations.

\textbf{Non-separability of breast and thorax.} 
Last but most importantly, the region of interest, i.e. the breast, obviously cannot be well separated from the rest of the thorax when considering 3D breast scans.
This is primarily because the chest wall separating the breast from the thorax cannot be captured using 3D surface scanning devices, but also due to the reason that no commonly accepted and \textit{exact} definition of the breast contour exists, see e.g. \cite{Lot21}. 
A statistical shape model built from 3D breast scans will therefore \textit{necessarily} capture also shape variations of the thorax, even after pose standardization.
In particular, these include morphological shape variations of the underlying chest wall and upper abdomen, but also those emerging from the arms, armpits, and shoulders due to improper pose standardization (also seen from Figure \ref{fig:database_samples}).
If not reduced to a minimum, this will cause the following unwanted effect.
Since breast and thorax shapes are tightly correlated in the subspace spanned by the model, the range of representable breast shapes is limited.
The reason is that a considerably large part of the subspace accounts for the unwanted shape variations of the surrounding areas.
Hence, the breast region should be decoupled from the thorax as much as possible in order to build an expressive and well-performing statistical shape model.
Due the absence of an exact definition of the breast contour, we assume the breast region to \textit{approximately} range from the lower breast pole to the second rib.

\subsection{Contributions}
\begin{figure}[t]
\includegraphics[]{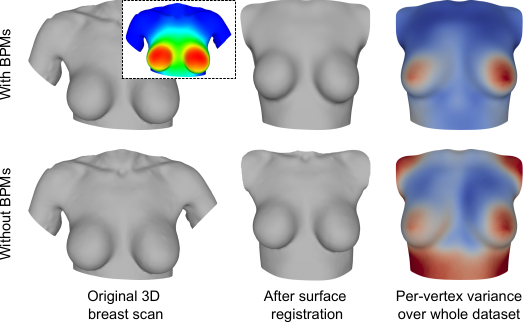}
\caption{The proposed concept of BPMs (top row) allows to minimize unwanted shape variations of the thorax by registering a template surface as precisely as possible \textit{inside} the breast region and only roughly outside. 
The breast region is defined by BPMs in a probabilistic manner (top left, red areas correspond to a high probability of belonging to the breast region). 
This simple yet effective strategy decouples the breast from the surrounding regions by reducing the variance outside the breast area. 
In the last column, the per-vertex variance over the whole dataset is visualized on the resulting mean shape. 
The regions showing the highest variance (red) are almost coincident with the breasts in our proposed BPM-based approach.
Contrary, without BPMs (bottom row), a lot of unwanted variance is present in the surrounding regions, especially around the arms, armpits, shoulders, and upper abdomen. 
This implies a strong coupling between breast and thorax in the final statistical shape model.}
\label{fig:overview_approach}
\end{figure}

As key contribution, this work presents the \textit{Regensburg Breast Shape Model} (RBSM) -- a 3D statistical shape model of the female breast.
In order to weaken the strong coupling between the breast and surrounding regions, we propose to minimize the \textit{variance} outside the breast region as much as possible.
To achieve this goal, a novel concept called \textit{breast probability masks} (BPMs) is introduced. 
A BPM assigns probabilities to each point of a 3D breast scan, telling how \textit{likely} it is that a particular point belongs to the breast region. 
Later, during pairwise registration, we use the BPMs to align the template to the target as accurately as possible \textit{inside} the breast region and only roughly outside. 
This way, only the most prominent and global shape variations outside the breast region will be recovered, effectively reducing the unwanted variance in these areas to a minimum. Figure \ref{fig:overview_approach} illustrates this idea.

To summarize, the contributions of this paper are three-fold:
\begin{itemize}
\item We introduce the \textit{Regensburg Breast Shape Model} (RBSM) -- an open-access 3D statistical shape model of the female breast built from 110 breast scans. It is available at \url{https://www.rbsm.re-mic.de/}.
\item We propose a fully automated, pairwise registration pipeline used to establish dense correspondence among our 3D breast scans. It uses \textit{breast probability masks} (BPMs) to decouple the breast from the surrounding regions as much as possible and requires only four landmarks to guide the registration process.
\item We present two exemplary applications demonstrating how the RBSM can be used for surgical outcome simulation and the prediction of a missing breast from the remaining one.
\end{itemize}
The remainder of this paper is organized as follows: Section \ref{sec:related_work} briefly reviews some related work. 
Section \ref{sec:methodology} describes the entire model building pipeline used to construct the RBSM. 
In particular, it formally introduces the notion of a BPM, followed by a detailed description of the proposed registration pipeline.
Section \ref{sec:evaluation} presents an extensive evaluation of the RBSM in terms of the common metrics compactness, generalization, and specificity. 
Using the RBSM, two exemplary applications are showcased in Section \ref{sec:applications}.
Finally, Section \ref{sec:discussion} discusses the results whereas Section \ref{sec:conclusion} concludes this article.

\section{Related work}
\label{sec:related_work}
In this section, we briefly summarize some related work concerning statistical shape models of the female breast, breast surgery simulation, and popular techniques for pairwise surface registration used within the context of statistical shape modeling in general.  
Note that we have limited our review to the 3D case. 

\textbf{Statistical shape models of the female breast.}
Literature about 3D statistical shape models of the female breast is sparse.
In the early work of Seo et al. \cite{Seo07}, a 3D statistical shape model was built from 28 breast scans with the goal of analyzing breast volume and surface measurements.
However, they assume symmetric breasts obtained by simply mirroring the right breast and did not make their model publicly available.
To date and to the best of our knowledge, this is the only work primarily addressing the construction of a statistical shape model from 3D breast scans, which is thus closest to our work.
Besides, Ruiz et al. \cite{Rui18} utilized a 3D statistical shape model built from 310 breast scans for the validation of a novel weighted regularized projection method used for 3D reconstruction.
As their focus did not lie on the construction of a well-performing statistical shape model of the breast, they did not provide detailed information about the registration method used to establish correspondence, the training data nor a comprehensive evaluation in terms of common metrics. 
Their model is also not publicly shared.

At last, few works exist attempting to construct a 3D statistical shape model from  Magnetic Resonance Imaging (MRI) data. 
Gallo et al. \cite{Gal10} applied \textit{Principal Component Analysis} to surface meshes extracted from 46 MRIs taken in prone position.
Further, Gallego and Martel \cite{Gal11} developed a statistical shape model from 415 semi-automatically segmented breast MRIs for model-based breast segmentation.

\textbf{Breast surgery simulation.}
Most of the existing methods for pre-operative breast surgery simulation are designed to simulate alloplastic, implant-based breast augmentation procedures, either for aesthetic reasons or after mastectomy as part of BRS.
Typically, those methods first generate a patient-specific, geometric representation of the breast (using tetrahedral meshes, for instance). 
Afterwards, the soft tissue deformation caused by implant insertion is simulated using a geometric and biomechanical model of the implant and breast, respectively.

As such, Roose et al. \cite{Roo06} used the tensor-mass model introduced by Cotin et al. \cite{Cot00}, a combination of classical finite element and mass-spring models, for implant-based breast augmentation planning.
De Heras Ciechomski et al. \cite{Cie12} proposed a web-based tool for breast augmentation planning which requires only 2D photographs and anthropometric measurements as input and allows the user to choose from a variety of different implants. 
Their method automatically reconstructs a 3D breast model and subsequently applies a tissue elastic model closely resembling the finite element model.
Georgii et al. \cite{Geo14} utilizes patient-specific finite element models generated from 3D breast scans. 
Combined with a novel mechanism called \textit{displacement template}, geometric implant models are no longer required, thus breaking up the coupling between implant and enclosing breast.

\begin{figure*}[t!]
\centering
\includegraphics{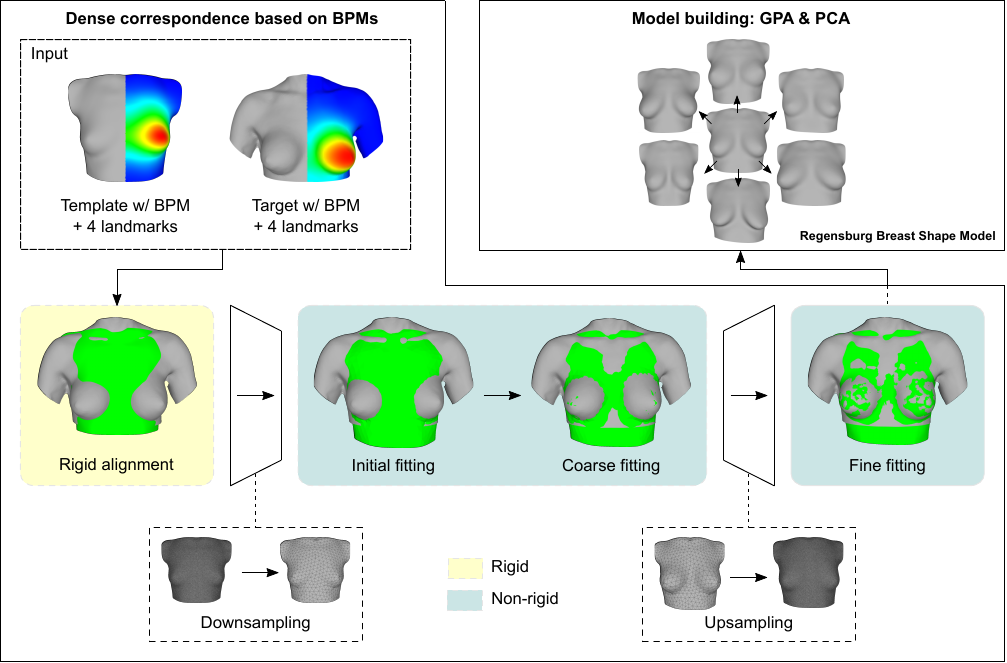}
\caption{An overview of the pipeline used to build the RBSM. 
We first establish dense correspondence by means of pairwise surface registration. 
Based on BPMs, our method starts by rigidly aligning the template to the target. 
Subsequently, a non-rigid alignment is applied in a hierarchical, multi-resolution fashion. 
In this step, BPMs are used to precisely recover only shape variations of the breast.
Finally, we perform classical \textit{Generalized Procrustes Analysis} (GPA) and \textit{Principal Component Analysis} (PCA) to build the model.}
\label{fig:overview_pipeline}
\end{figure*}

Besides the simulation of breast augmentation procedures using implants, some methods exist especially addressing the simulation of BRS \textit{without} implant insertion, for example by means of autologous fat tissue.
Based on Pascal's principle and volume conservation, Costa and Balaniuk \cite{Cos01} developed a novel approach for real-time physically based simulation of deformable objects, called \textit{Long Elements Method} (LEM).
An extension of LEM, known as the \textit{Radial Elements Method} \cite{Bal03}, was later used by Balaniuk et al. \cite{Bal06} for cosmetic and reconstructive breast surgery simulation.
Williams et al. \cite{Wil03} employ a finite element approach incorporating the Mooney-Rivlin hyperelastic material model for the realistic simulation of soft tissue to simulate \textit{transverse rectus abdominis myocutaneous} flap breast reconstruction.

Although the majority of the works on state-of-the-art breast surgery simulation utilize biomechanical models to predict the breast shape after surgery, an early attempt by Kim et al. \cite{Kim08} employs an example-based approach. 
Using a \textit{sparse} set of 3D feature point pairs collected on 30 patients before and after surgery as training database, they developed a linear combination model to predict the surgical outcome.

For most of the existing methods, however, various authors highlighted the disagreement between simulated and actual outcomes \cite{Cru15,Roo06} or only partially satisfying results for certain types of breast shapes \cite{Vor17}.

\textbf{Pairwise surface registration.}
During the last few decades, countless algorithms were proposed tackling the problem of (pairwise) non-rigid surface registration.
To date, however, none of them were used for 3D breast scan registration.

One of the most widely used methods is the \textit{Optimal Step Non-rigid Iterative Closest Point} (NICP) framework proposed by Amberg et al. \cite{Amb07} and based on the work of Allen et al. \cite{All03}. 
NICP is the only method already used in the female breast shape domain to reconstruct a 3D breast model from a sequence of depth images \cite{Lac19}.
Moreover, NICP was employed to construct the BFM and LSFM.
A second class of non-rigid registration methods is based on splines, such as \textit{Thin Plate Splines} (TPS) or B-splines.
Pioneered by Bookstein \cite{Boo89}, TPS were utilized by Paulsen et al. \cite{Pau02} to construct a statistical shape model of the human ear canal.
B-splines are extensively used in the \textit{free-form deformations} framework and, among others, used for the creation of shape models of the human heart \cite{Ord07}.
A third, recently introduced framework is based on \textit{Gaussian Process Morphable Models} (GPMMs) introduced by Lüthi et al. \cite{Lue18}.
GPMMs are statistical shape models themselves generalizing classical point-based models as proposed by Cootes et al. \cite{Coo95}.
By means of GPMMs, expected deformations can be modeled using analytic covariance functions and later used as a \textit{prior} for non-rigid surface registration, thus effectively reducing the search space. 
This approach was successfully used by Gerig et al. \cite{Ger17} to build an improved version of the BFM including facial expressions. 
Lastly, pairwise non-rigid surface registration algorithms are built upon the \textit{as-rigid-as-possible} (ARAP) real-time mesh deformation framework proposed by Sorkine and Alexa \cite{Sor07}.
This method was successfully transferred to the registration domain, yielding to similar methods constraining deformations to be \textit{as conformal as possible} \cite{Yos14} or \textit{as similar as possible} \cite{Jia17,Yam13}. 
Recently, a variant of ARAP was utilized by Dai et al. \cite{Dai20} to built a shape model of the full human head. 

\section{Methodology}
\label{sec:methodology}
This section describes the entire pipeline used to build the RBSM. 
As outlined in Figure \ref{fig:overview_pipeline}, we start by establishing dense correspondence among our training data.
Based on a novel concept called \textit{breast probability masks} (Section \ref{subsec:bpms}), this is achieved by means of a fully automated, pairwise registration pipeline as proposed in Section \ref{subsec:registration}.
Finally, we follow the typical workflow used to build a point-based statistical shape model by applying \textit{Generalized Procrustes Analysis} and \textit{Principal Component Analysis} to the registered data set (briefly summarized in Section \ref{subsec:model_building}).

In what follows, 3D breast scans are represented using triangular surface meshes.
A triangle mesh $\mathcal{M}=(V,E,\mathcal{P})$ is fully specified by a set of $n$ vertices $V\subset\mathbb{N}$, edges $E\subset V\times V$, and an embedding $\mathcal{P}=\{\mathbf{p}_1,\mathbf{p}_2\ldots,\mathbf{p}_n\}\subset\mathbb{R}^3$. 
Sometimes, however, instead of arranging points $\mathbf{p}_i$ into a set, it is more convenient to use a matrix notation $\mathbf{P}=(\mathbf{p}_1,\mathbf{p}_2,\ldots,\mathbf{p}_n)^\top\in\mathbb{R}^{n\times 3}$.
Hence, we will denote a triangle mesh either as $\mathcal{M}=(V,E,\mathcal{P})$ or equivalently as $\mathcal{M}=(V,E,\mathbf{P})$.

\subsection{Breast probability masks}
\label{subsec:bpms}
Given a 3D breast scan represented as triangle mesh $\mathcal{M}=(V,E,\mathcal{P})$, we call
\begin{equation}
p_\mathcal{M}:\mathcal{P}\longrightarrow(0,1]
\end{equation}
a \textit{breast probability mask} (BPM). Technically, a BPM is a scalar field defined over $\mathcal{M}$ assigning each point $\mathbf{p}_i$ of a 3D breast scan a probability $p_\mathcal{M}(\mathbf{p}_i)$ telling how \textit{likely} it is that $\mathbf{p}_i$ belongs to the breast region.

\begin{figure}[t!]
\includegraphics[]{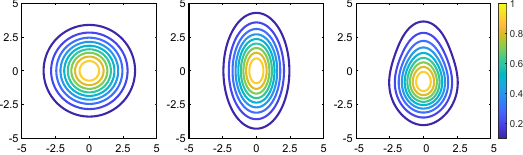}
\caption{From left to right: comparison between RBF, EBF, and a sum of two EBFs, illustrated as contour plots. 
While a simple RBF or EBF is not able to accurately mimic the typical teardrop shape of the breast, a sum of two EBFs comes close.}
\label{fig:comparison_bpms}
\end{figure}

\textbf{Concrete mapping.} As a concrete mapping for $p_\mathcal{M}$ we propose to use a normalized sum of \textit{elliptical basis functions} (EBFs), centered at the nipples.
We use EBFs instead of ordinary \textit{radial basis functions} (RBFs) because we found that they better capture the natural teardrop shape of the breast (see Figure \ref{fig:comparison_bpms} for a comparison between RBFs and EBFs).
Technically, EBFs are a generalization of RBFs using the Mahalanobis distance instead of an ordinary vector norm.
Formally, an EBF $\phi:[0,\infty)\longrightarrow\mathbb{R}$ centered at a point $\mathbf{c}\in\mathbb{R}^n$ is of the form $\phi(\mathbf{x})=\phi\left(d_M\left(\mathbf{x},\mathbf{c}\right)\right)$. 
Here, $d_M$ is the Mahalanobis distance, defined as
\begin{equation}
d_M(\mathbf{x},\mathbf{c})\coloneqq\sqrt{\left(\mathbf{x}-\mathbf{c}\right)^\top\mathbf{S}^{-1}\left(\mathbf{x}-\mathbf{c}\right)}\,,
\end{equation}
where $\mathbf{S}\in\mathbb{R}^{n\times n}$ is a symmetric positive definite matrix, also called \textit{covariance matrix}.
To stress that the Mahalanobis distance depends on $\mathbf{S}$, we write $d_M(\mathbf{x},\mathbf{c};\mathbf{S})$ in the following.

Now, in order to define a concrete BPM using EBFs, let $\mathbf{p}_\text{N}^\tau\in\mathcal{P}$ denote the position of the left (L) and right (R) nipple, respectively, and $\tau\in\{\text{L, R}\}$.
We first construct two individual probability masks for the left and the right breast, given as
\begin{equation}
p_\mathcal{M}^\tau(\mathbf{p}_i)=\phi\left(d_M\left(\mathbf{p}_i,\mathbf{p}_\text{N}^\tau;\mathbf{S}_\tau\right)\right).
\end{equation}
Hereby, we define $\phi:[0,\infty)\longrightarrow(0,1]$ as 
\begin{equation}
\phi(x)=\exp\left(-x^2\right).
\end{equation}
Finally, the BPM for a whole 3D breast scan is given as the normalized sum
\begin{equation}
p_\mathcal{M}(\mathbf{p}_i)=\frac{1}{4}\left(p_\mathcal{M}^\text{L}(\mathbf{p}_i)+\hat{p}_\mathcal{M}^\text{L}(\mathbf{p}_i)+p_\mathcal{M}^\text{R}(\mathbf{p}_i)+\hat{p}_\mathcal{M}^\text{R}(\mathbf{p}_i)\right),
\end{equation}
where 
\begin{equation}
\hat{p}_\mathcal{M}^\tau(\mathbf{p}_i)=\phi\left(d_M\left(\mathbf{p}_i,\mathbf{\hat{p}}_\text{N}^\tau;\mathbf{\hat{S}}_\tau\right)\right)
\end{equation}
are shifted BPMs of the left and right breast added to better mimic the teardrop shape, and $\mathbf{\hat{p}}_\text{N}^\tau=\mathbf{p}_\text{N}^\tau+\mathbf{t}_\tau$ with translation vectors $\mathbf{t}_\tau\in\mathbb{R}^3$.

\textbf{Parameter selection.} In order to fully define a BPM, appropriate matrices $\mathbf{S}_\tau,\mathbf{\hat{S}}_\tau\in\mathbb{R}^{3\times 3}$ and translation vectors $\mathbf{t}_\tau\in\mathbb{R}^3$ need to be chosen first.
As such, a total of 30 values are required to be properly determined (six for each $\mathbf{S}_\tau$ and $\mathbf{\hat{S}}_\tau$, and three for each $\mathbf{t}_\tau$). 
To simplify that task, we assume diagonal covariance matrices and utilize previously expert-marked landmarks on the 3D breast scans.
Specifically, denote the landmark points shown in Figure \ref{fig:anatomy} as $\mathbf{p}_\text{SN},\mathbf{p}_\text{XI}\in\mathcal{P}$ for sternal notch and xiphoid, and $\mathbf{p}_\text{LaBP}^\tau,\mathbf{p}_\text{LBP}^\tau\in\mathcal{P}$ for left and right lateral and lower breast pole, respectively.
We then define
\begin{equation}
\begin{split}
\mathbf{S}_\tau&=
\!\begin{multlined}[t][60mm]
\frac{1}{2}\diag\Bigl(d_G\left(\mathbf{p}^\tau_\text{LaBP},\mathbf{p}^\tau_\text{N}\right)+d_G\left(\mathbf{p}^\tau_\text{N},\mathbf{p}_\text{XI}\right),\\
d_G\left(\mathbf{p}^\tau_\text{N},\mathbf{p}^\tau_\text{LBP}\right), d_G\left(\mathbf{p}^\tau_\text{LaBP},\mathbf{p}^\tau_\text{N}\right)\Bigr)\,,
\end{multlined}\\
\mathbf{\hat{S}}_\tau&=
\!\begin{multlined}[t][60mm]
\frac{1}{2}\diag\Bigl(d_G\left(\mathbf{p}^\tau_\text{LaBP},\mathbf{p}^\tau_\text{N}\right)+d_G\left(\mathbf{p}^\tau_\text{N},\mathbf{p}_\text{XI}\right),\\
d_G\left(\mathbf{p}^\tau_\text{N},\mathbf{p}_\text{SN}\right),d_G\left(\mathbf{p}^\tau_\text{LaBP},\mathbf{p}^\tau_\text{N}\right)\Bigr)\,,
\end{multlined}\\
\mathbf{t}_\tau&=\mathbf{p}^\tau_\text{N}+\frac{1}{5}\left(0,d_G\left(\mathbf{p}^\tau_\text{N},\mathbf{p}_\text{SN}\right),0\right),
\end{split}
\end{equation}
where $d_G$ denotes the Geodesic distance between two points on the surface mesh.
Note that $\mathbf{S}_\tau$ and $\mathbf{\hat{S}}_\tau$ differ only in the second diagonal element.

\subsection{Registration of 3D breast scans}
\label{subsec:registration}
Following Figure \ref{fig:overview_pipeline}, the proposed pairwise registration pipeline is mainly composed of rigid alignment (Section \ref{subsubsec:rigid}) and non-rigid alignment (Section \ref{subsubsec:non-rigid}). 
To speed up convergence, the latter is carried out in a hierarchical, multi-resolution fashion (Section \ref{subsubsec:multi_res_fitting}).

Both phases make extensive use of BPMs in order to align a template surface $\mathcal{S}=(V,E,\mathbf{P})$ to a target $\mathcal{T}$ as accurately as possible \textit{inside} the breast region and only roughly outside, effectively decoupling the breast from the rest of the thorax by reducing the \textit{variance} outside the breast region to a minimum.
This is justified as the covariance $\cov(x,y)$ becomes smaller if $\var(x)$ or $\var(y)$ is lowered, following from the well known fact that $\left|\cov(x,y)\right|\leq\sqrt{\var(x)}\sqrt{\var(y)}$ (which holds via the Cauchy–Schwarz inequality).

Finally, note that the target surface $\mathcal{T}$ can be given in any representation that allows for closest point search.
We use a triangular surface mesh but write $\mathcal{T}\subset\mathbb{R}^3$ for the sake of notational simplicity.

\subsubsection{Rigid alignment}
\label{subsubsec:rigid}
The overall goal of the rigid alignment is to move the template as close as possible to the \textit{rigid part} of the target, which we define as \textit{the thorax without the breast}.
In particular, we expect that the thoraxes of two subjects without the breast region can be sufficiently well aligned if we assume the breast to be the only part of the thorax that deforms non-rigidly.
Based on this assumption, the absence of suitable landmarks, and due to the fact that our initial 3D breast scans are already reasonably well aligned (see Section \ref{subsec:data}) we propose a modified version of the \textit{ Iterative Closest Point} (ICP) algorithm, originally introduced by Besl and McKay \cite{Bes92}.

Essentially, compared to the standard version of the ICP algorithm, our modified version differs in the following three aspects: (i) a scaling factor is added to the rigid transformation effectively allowing for Euclidean similarity transformations \cite{Du07,Zin05}. 
Secondly, (ii) to ensure that only the rigid parts of the 3D breast scans are used for alignment, correspondences, where both points have a high probability belonging to the breast region, are discarded.
This is implemented by thresholding the template and target BPMs.
Finally, (iii) rotations are restricted to the $x$-axis corresponding to the transversal plane.
Rotations around the $y$- and $z$-axis (sagittal and coronal plane), possibly introduced due to severe overweight in conjunction with an uneven distribution of abdominal fat could destroy the initial alignment and lead to misalignment.
In any case, asymmetries introduced due to the thorax should \textit{not} affect the rigid alignment of the template.

\subsubsection{Non-rigid alignment}
\label{subsubsec:non-rigid}
Given the rigidly aligned template $\mathcal{S}=(V,E,\mathbf{P})$, the goal of the non-rigid alignment is to gradually deform $\mathcal{S}$ into a new surface $\mathcal{S}'=(V,E,\mathbf{P}')$ with identical topology such that $\mathcal{S}'$ is as close as possible to the target $\mathcal{T}$ \textit{inside} the breast region.
Following various authors including Jiang et al. \cite{Jia17} and Yamazaki et al. \cite{Yam13}, we formulate our non-rigid registration problem using the following non-linear energy functional
\begin{equation}
F\left(\mathbf{P}'\right)=F_D\left(\mathbf{P}'\right)+\alpha F_R\left(\mathbf{P}'\right)+\beta F_L\left(\mathbf{P}'\right),
\label{eq:non_rigid_cost}
\end{equation}
where $F_D$ is a distance term used to penalize the point-to-point distance between the template and target surface, $F_R$ is a regularization term constraining deformations \textit{as similar as possible}, and $F_L$ constitutes a landmark term ensuring certain points to be matched.
$\alpha,\beta\geq 0$ are weights controlling the individual contribution of each term to the cost function.
Minimizing $F$ finally leads to the new points $\mathbf{P}'$ of the deformed template surface $\mathcal{S}'$, i.e.
\begin{equation}
\mathbf{P}'=\argmin_{\mathbf{P'}\in\mathbb{R}^{n\times 3}}F(\mathbf{P'}).
\label{eq:non_rigid_opt_problem}
\end{equation}
Adapting the strategy proposed by Allen et al. \cite{All03}, instead of computing (\ref{eq:non_rigid_opt_problem}) only once, we minimize $F$ several times but each time lowering the regularization weight $\alpha$ in (\ref{eq:non_rigid_cost}).
As later demonstrated by Amberg et al. \cite{Amb07}, this strategy is able to recover the whole range of global and local non-rigid deformations efficiently.
Following various authors \cite{Jia17,Sor07}, the optimization problem in (\ref{eq:non_rigid_opt_problem}) is solved using an alternating minimization (AM) approach as briefly summarized in \ref{app:solve_am}.

\textbf{Distance term.}
The distance term $F_D$ is used to attract the template $\mathcal{S}$ to the target $\mathcal{T}$.
Assuming fixed correspondences between both surfaces, i.e. $\left\{(\mathbf{p}_1,\mathbf{q}_1),(\mathbf{p}_2,\mathbf{q}_2),\ldots,(\mathbf{p}_n,\mathbf{q}_n)\right\}$ with $\mathbf{q}_i\in\mathcal{T}$ being the closest point to $\mathbf{p}_i$, the distance term can be written as
\begin{equation}
F_D(\mathbf{P}')=\frac{1}{2}\left\Vert\mathbf{C}\left(\mathbf{P}'-\mathbf{Q}\right)\right\Vert^2_F,
\end{equation}
where $\mathbf{C}\coloneqq\text{diag}(c_1,c_2,\ldots,c_n)$, $c_i\geq 0$ for all $i\in\left\{1,2,\ldots,n\right\}$ are weights used to quantify the reliability of a match, and $\mathbf{Q}\coloneqq\left(\mathbf{q}_1,\mathbf{q}_2,\ldots,\mathbf{q}_n\right)^\top\in\mathbb{R}^{n\times 3}$.
Using the BPMs $p_\mathcal{S}$ and $p_\mathcal{T}$ of the template and target, we set
\begin{equation}
c_i=\frac{p_\mathcal{S}(\mathbf{p}_i)+p_\mathcal{T}(\mathbf{q}_i)}{2}.
\end{equation}
This way, correspondences $(\mathbf{p}_i,\mathbf{q}_i)$ mapping from one breast region to the other have a greater impact on the overall distance term as $c_i\in(0,1]$ becomes large in this case.
Conversely, the influence tends to zero if $c_i\rightarrow 0$, i.e. if both points are less likely to belong to the breast region.
As such, the deformation of points $\mathbf{p}_i$ on the template with a small value for $c_i$ is mainly controlled by the regularization term, as previously described by Allen et al. \cite{All03}.

\textbf{Regularization term.}
The regularization term $F_R$ should prevent the template surface from shearing and distortion while simultaneously ensuring structure preservation and smooth deformations.
To do so, we adapt the \textit{consistent as-similar-as-possible} (CASAP) regularization technique in which deformations are constrained to be \textit{locally} as similar as possible \cite{Jia17,Yam13}.
Specifically, given a local neighborhood $E_i\subset E$ around each point $\mathbf{p}_i$, the template surface is only allowed to move in terms of an Euclidean similarity transformation
\begin{equation}
\mathbf{p}'_j-\mathbf{p}'_k=s_i\mathbf{R}_i\left(\mathbf{p}_j-\mathbf{p}_k\right)\qquad\forall(j,k)\in E_i,
\label{eq:asap}
\end{equation}
where $s_i>0$ is a scaling factor and $\mathbf{R}_i\in\SO(3)$ a rotation matrix.
Following Chao et al. \cite{Cha10}, we define $E_i$ as \textit{the set containing all (directed) edges of triangles incident to} $\mathbf{p}_i$, also known as \textit{spokes-and-rims}.
Finally, our CASAP regularization term reads
\begin{multline}
F_R(\mathbf{P}')=\frac{1}{2}\sum_{i=1}^n w_i\left[\sum_{(j,k)\in E_i}w_{jk}\left\Vert\left(\mathbf{p}'_j-\mathbf{p}'_k\right)-s_i\mathbf{R}_i\left(\mathbf{p}_j-\mathbf{p}_k\right)\right\Vert^2_2+\right.\\
\left.\lambda\sum_{l\in N_i}w_{il}\left\Vert\mathbf{R}_i-\mathbf{R}_l\right\Vert^2_F\right]\,,
\label{eq:casap_regularization}
\end{multline}
where weights $w_i>0$ are added to individually control the amount of regularization for each particular point.
As mentioned above, since the deformation of points $\mathbf{p}_i$ with a small value for $c_i$ is mainly controlled by the regularization term, we define
\begin{equation}
w_i=\frac{1}{(h-1)c_i+1}\qquad\text{with}\qquad\frac{1}{h}\leq w_i<1
\end{equation}
for all $i\in\{1,2,\ldots,n\}$ and $h\in\mathbb{N}^+$ (we used $h=2$ throughout this paper).
As seen, this strategy keeps points $\mathbf{p}_i$ of the template relatively stiff if (i) $\mathbf{p}_i$ has a low probability belonging to the breast region and (ii) if the corresponding point on the target is also not likely to be part of the breast region (because $w_i\rightarrow 1$ if $c_i\rightarrow 0$), thus effectively preventing the template from adapting too close to the target outside the breast region.
Lastly, $N_i\subset V$ in (\ref{eq:casap_regularization}) denotes the one-ring neighborhood of the $i$-th point and $w_{jk}\in\mathbb{R}$ are the popular cotangent weights, see e.g. \cite{Bot10}.
$\lambda\geq 0$ is usually set to $0.02A$, where $A\geq 0$ is the total surface area of $\mathcal{S}$ \cite{Lev14}.

\textbf{Landmark term.}
The goal of the landmark term $F_L$ is to keep certain positions (i.e. landmarks) fix during the registration process.
Let $I\subset\mathbb{N}$ be an index set containing the indices of the $m$ landmarks specified on the template surface $\mathcal{S}$.
Define a matrix $\mathbf{D}\in\mathbb{R}^{m\times n}$ as
\begin{equation}
\mathbf{D}=(d_{ij}):=
\begin{cases}
1, & \text{if }j\in I,\\
0, & \text{otherwise}
\end{cases}
\end{equation}
for $i=1,2,\ldots,m$ and $j=1,2,\ldots,n$.
Next, denote the corresponding landmarks on the target surface by $\{\mathbf{q}_1,\mathbf{q}_2,\ldots,\mathbf{q}_m\}\subset\mathcal{T}$. 
Then, the landmark term is defined as
\begin{equation}
F_L(\mathbf{P}')=\frac{1}{2}\left\Vert\mathbf{DP}'-\mathbf{Q}_L\right\Vert^2_F,
\end{equation}
where $\mathbf{Q}_L\coloneqq\left(\mathbf{q}_1,\mathbf{q}_2,\ldots,\mathbf{q}_m\right)^\top\in\mathbb{R}^{m\times 3}$.

\subsubsection{Multi-resolution fitting strategy}
\label{subsubsec:multi_res_fitting}
Following common practice and to speed up convergence, instead of applying the previously described non-rigid alignment only once, we employ a hierarchical, multi-resolution fitting strategy composed of initial fitting, coarse fitting, and fine fitting (see also Figure \ref{fig:overview_pipeline}).

\textbf{Initial fitting.}
Having a low-resolution instance of the rigidly aligned template at hand, the goal of the initial fitting is to roughly adapt the coarse template to the key features (i.e. landmarks) of the target.
To do so, we strictly prioritize the landmark constraints and do not use BPMs in this phase.

\textbf{Coarse fitting.}
In this step, the initially fitted low-resolution template is gradually deformed towards the target.

\textbf{Upsampling.}
Next, the deformation obtained from the previous step is applied to the original, full-resolution template. 
This is achieved using a concept called \textit{Embedded Deformation}, introduced by Sumner et al. \cite{Sum07}.
In essence, the deformation of the coarse template obtained from the previous step is transferred to the template by linearly interpolating the transformation at each point.

\textbf{Fine fitting.}
Lastly, the upsampled template is fitted to the target again to produce the final result. 

\subsection{Model building}
\label{subsec:model_building}
Once the data set is brought into correspondence, we follow the typical workflow used to build a classical point-based statistical shape model as proposed by Cootes et al. \cite{Coo95}.
For notational simplicity, instead of stacking points $\mathcal{P}$ of a triangular mesh $\mathcal{M}=(V,E,\mathcal{P})$ into a matrix $\mathbf{P}\in\mathbb{R}^{n\times 3}$ as before, we use a vectorized representation, denoted as $\mathbf{x}=\vecz(\mathbf{P})\in\mathbb{R}^{3n}$ in the following.

Briefly, given a set of $k$ breast scans $\{\mathbf{x}_1,\mathbf{x}_2,\ldots,\mathbf{x}_k\}\subset\mathbb{R}^{3n}$ in correspondence, we first perform \textit{Generalized Procrustes Analysis} (GPA) as introduced by Gower \cite{Gow75}.
GPA iteratively aligns the objects to the arithmetic mean $\mathbf{\bar{x}}\in\mathbb{R}^{3n}$ (successively estimated from the data) by using an Euclidean similarity transformation, effectively transforming the objects into the shape space.
Secondly, \textit{Principal Component Analysis} (PCA) is carried out on the Procrustes-aligned shapes.
Let $\{\lambda_1,\lambda_2,\ldots,\lambda_q\}\subset\mathbb{R}^+$ be the $q<k$ non-zero eigenvalues (also called \textit{Principal Components} (PCs) in this context) of the empirical covariance matrix calculated from the data and sorted in a descending order.
Denote the corresponding eigenvectors as $\{\mathbf{u}_1,\mathbf{u}_2,\ldots,\mathbf{u}_q\}\subset\mathbb{R}^{3n}$.
Then, a statistical shape model can be interpreted as a linear function $M:\mathbb{R}^q\longrightarrow\mathbb{R}^{3n}$ defined as
\begin{equation}
M(\boldsymbol\alpha)=\mathbf{\bar{x}}+\mathbf{U}\boldsymbol\alpha,
\end{equation}
where $\mathbf{U}\coloneqq(\mathbf{u}_1,\mathbf{u}_2,\ldots,\mathbf{u}_q)\in\mathbb{R}^{3n\times q}$.
To ensure plausibility of the newly generated shapes, $\alpha_i$ is usually restricted to $\left|\alpha_i\right|\leq 3\sqrt{\lambda_i}$ for all $i\in\{1,2,\ldots,q\}$.
If a (possibly unseen) shape $\mathbf{x}'\in\mathbb{R}^{3n}$ is in correspondence with the model and properly aligned, it can be reconstructed from $M$ in a least-squares sense by using
\begin{equation}
\boldsymbol\alpha^*=\mathbf{U}^{-1}\left(\mathbf{x}' - \mathbf{\bar{x}}\right)
\end{equation}
as the model parameters, i.e. $\mathbf{x}'\approx M(\boldsymbol\alpha^*)$.
The number $q<k$ of retained PCs is chosen so that the model typically represents a fixed proportion of the total variance, e.g. 98\%.

\begin{table}[b!]
\caption{An overview of the 3D breast scan database used to build the RBSM. It includes 110 textured 3D breast scans.}
\label{tab:overview_database} 
\vspace{2mm}
\centering
\begin{tabular}{p{22.1mm} l l}
\hline
& Mean ($\pm$ SD) & Range [min -- max]\\
\hline
Age [years] & 40.78 ($\pm$ 14.30) & 18.00 -- 83.00\\
BMI [kg/m$^2$] & 23.66 ($\pm$ 3.57) & 16.90 -- 38.27\\
\hline
\multicolumn{3}{l}{Breast volume [cc]} \\
\hspace{3mm} Left & 477.23 ($\pm$ 242.94) & 70.60 -- 1,258.90\\
\hspace{3mm} Right & 481.45 ($\pm$ 240.27) & 80.00 -- 1,609.30\\
\hspace{3mm} Difference & \hspace{.95mm} 80.12 ($\pm$ 81.34) & \hspace{.95mm} 0.30 -- 367.40\\
\hline
\end{tabular}
\end{table}

\section{Evaluation}
\label{sec:evaluation}
Based on our 3D breast scan database which we present in Section \ref{subsec:data}, several experiments were conducted in order to evaluate the proposed statistical shape model (Section \ref{subsec:exp_res}).

\subsection{Data}
\label{subsec:data}
Our database consists of 110 textured 3D breast scans collected at our institution (St. Josef Hospital Regensburg) using the portable Vectra H2$^{\text{TM}}$ scanning system (Canfield Scientific, New Jersey, USA).
The H2 system employs photogrammetry to reconstruct a 3D surface mesh from a series of 2D images with a resolution of 3.5 mm and a maximal capture volume of $70\times 41\times 40$ cm$^3$ (width $\times$ height $\times$ depth).
In our case, in total three photos were taken of each subject: one from a frontal view and two from a lateral view ($\pm 45^\circ$).
Note that due to the standardized distance and angles from which the photos are taken, the reconstructed 3D breast scans are already reasonably well aligned and consistently oriented.

\begin{figure*}[t!]
\centering
\includegraphics{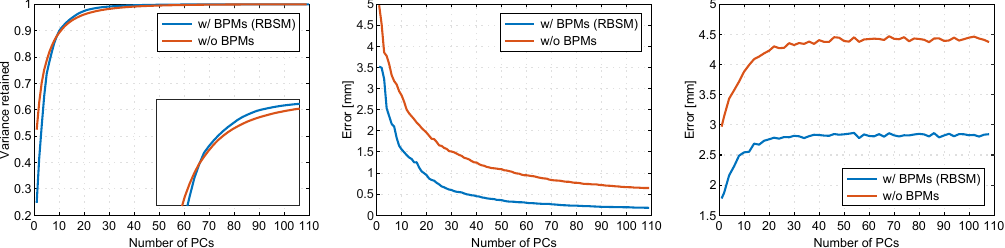}
\caption{From left to right: compactness, generalization, and specificity of statistical shape models built from 3D breast scans registered either with (i.e. the RBSM) or without using BPMs during registration.}
\label{fig:exp1_model}
\end{figure*}

A previously developed standardized scanning protocol \cite{Har20} was used to ensure a common pose and same imaging conditions for all participants.
In brief, all subjects were asked to stand in an upright posture and abduct both arms by an angle of $45^\circ$ (a telescopic stick was used to support the subjects in holding their arms fixed at the specified angle).
This posture produces a natural-looking breast shape and allows to capture the whole breast as isolated as possible.
Moreover, it is fast and easy to adopt for all participants regardless of age, body weight, or medical history.

A compact overview of some key parameters in our database is given in Table \ref{tab:overview_database}. 
In addition, 62 out of 110 participants have at least one ptotic breast.
Of these, 15 women show different ptosis grades on the left and right breast.
72 out of 110 participants received some kind of breast surgery such as alloplastic or autologous breast reconstruction or augmentation.

\subsection{Experiments and results}
\label{subsec:exp_res}
Following common practice, we evaluate our statistical shape model using the well-known metrics compactness, generalization, and specificity as proposed by Styner et al. \cite{Sty03}.
Registration results are assessed by the classical distance-based \textit{mean squared error} (MSE) between deformed template and target.
In addition, the \textit{angle distortion} between the deformed template $\mathcal{S}'$ and the original template $\mathcal{S}$ is determined in order to quantify the amount of shearing introduced due to non-rigid registration. 
This is achieved by simply averaging the absolute deviation between the inner-triangle angles of $\mathcal{S}$ and $\mathcal{S}'$ over all triangles \cite{Wan16}.

For all experiments, the same template $\mathcal{S}$ was used during registration.
In particular, the 3D breast scan of a healthy, non-operated subject was chosen and subsequently mirrored along the sagittal plane to produce a perfectly symmetrical template. 
After isotropic remeshing and Laplacian smoothing, a regular surface mesh with an average edge length of 2.05 mm was obtained.
It consists of 30,924 vertices.
The coarse, low-resolution version of our template was created using mesh simplification techniques and includes 646 vertices. All relevant parameters used for registration are listed in \ref{app:params}.

The complete registration pipeline was implemented in C++.
Statistical shape modeling was performed with an open-source framework called Scalismo (\url{https://scalismo.org/}), implemented in Scala and based on Statismo \cite{Lue12}.

\begin{figure}[t]
\includegraphics[]{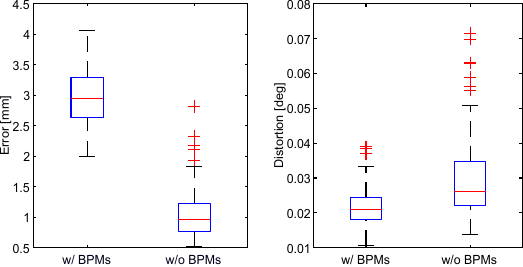}
\caption{MSE (left) and angle distortion (right), calculated from 3D breast scans registered with and without BPMs. }
\label{fig:exp1_registration}
\end{figure}
\begin{figure}[b!]
\includegraphics{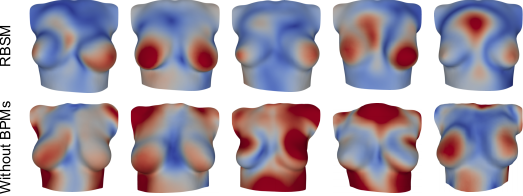}
\caption{Comparison between five random samples from the RBSM (top row, generated from 3D breast scans registered using BPMs) and five random samples from a model constructed without using BPMs (bottom row). 
The samples are colored according to the distance from the mean $\mathbf{\bar{x}}$. 
As such, the red areas underwent a high deformation, whereas blue regions are rather stiff. For samples drawn from the RBSM, the regions showing the highest variation are almost always the breast region, indicating a solid decoupling between breast shape and thorax.}
\label{fig:exp1_random_samples}
\end{figure}

\subsubsection{On the effect of BPMs}
The first experiment should evaluate our novel BPM-based registration technique, its influence on the resulting statistical shape models, and, in particular, whether or not BPMs are able to weaken the strong coupling between the breast and surrounding regions.
As such, we compare two different models: the first one is constructed from 3D breast scans brought into correspondence using the proposed approach based on BPMs (i.e. the RBSM). 
The second model is built from 3D breast scans registered without using BPMs.

Figure \ref{fig:exp1_model} shows the evaluation metrics of the resulting statistical shape models. 
As seen, although the model built without BPMs is more compact than with BPMs (i.e. the RBSM) when considering only the very first PCs, it is also the model that generalizes worse.
Specifically, when using all 109 available PCs a generalization error of 0.65 mm is reported.
For comparison, the RBSM shows a generalization error of only 0.17 mm when using the same number of PCs. 
Finally, the RBSM constantly achieves the lower specificity error of about 2.8 mm if more than 30 PCs are used.

\begin{figure*}[t!]
\includegraphics[width=\textwidth]{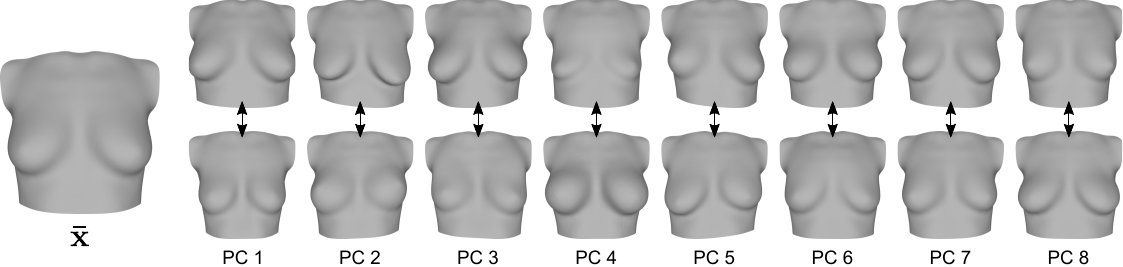}
\caption{The first eight principal modes of variation from the RBSM, visualized by either adding (top row) or subtracting (bottom row) $3\sqrt{\lambda_i}\mathbf{u}_i$ from the mean $\mathbf{\bar{x}}$, displayed on the left. 
Together, they represent about 85\% of the total variance present in the dataset.}
\label{fig:modes_rbsm}
\end{figure*}

\begin{figure*}[b]
\centering
\includegraphics{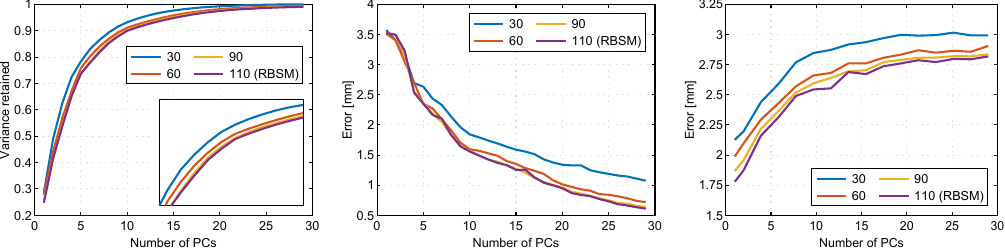}
\caption{From left to right: compactness, generalization, and specificity of statistical shape models built from randomly sampled subsets of the training data.}
\label{fig:exp2_model}
\end{figure*}

Figure \ref{fig:exp1_registration} summarizes the registration results, quantified in terms of MSE and angle distortion.
As expected, registration without using BPMs clearly achieves the lower MSE of 1.05 mm as the goal here is to register the template to the target as close as possible.
For comparison, the MSE achieved when using BPMs is 2.98 mm.
However, it is important to remember that the goal of the BPM-based registration is to align the template as precisely as possible to the target only \textit{inside} the breast region.
Therefore, the overall MSE between template and target might not reflect the actual registration quality well (see e.g. Figure \ref{fig:overview_approach}).
In terms of angle distortion, only a small difference is noticeable between registration with and without BPMs, respectively.
In particular, registration without BPMs caused the angle distortion to increase by only 0.008 degrees on average.

In order to further investigate to which extend BPM-based registration is able to weaken the strong coupling between breast region and thorax, Figure \ref{fig:exp1_random_samples} shows some random samples from the RBSM and the model built from 3D breast scans registered without BPMs.
The samples are colored according to their distance to the mean shape, effectively providing a \textit{measure of variation}.
Interestingly, the areas of the highest variation in the samples drawn from the RBSM are almost always located at the breasts.
The surrounding regions are rather stiff, indicating low variance and a quite well decoupling between breast and thorax.
Hence, the RBSM is able to produce a variety of breast shapes without altering the whole thorax too much, also reflected in the principal modes of variation shown in Figure \ref{fig:modes_rbsm}. 
Contrary, the area of the highest variation in the samples generated from the model built without BPMs is oftentimes \textit{not} the breast region.

\subsubsection{How much data is needed?}
An often arising question in the context of statistical shape modeling concerns the amount of training data needed to build a reasonably well-performing model.
Generally, this question very much depends on the amount of variability samples from a particular class of objects are expected to show.
As a rule of thumb, a good training set should always reflect the whole bandwidth of possible variations likely to occur within a target population.
The goal of this experiment is to test how much data is needed to build a well-performing statistical shape model of the female breast.
To do so, we randomly sample 30, 60, and 90 breast scans from our database and subsequently compare the resulting models with the model containing all 110 breast scans (i.e. the RBSM).
To avoid sampling bias, the whole procedure was repeated three times, and results were averaged.

Figure \ref{fig:exp2_model} shows the results in terms of compactness, generalization, and specificity.
Regarding compactness, it can be clearly seen that the model built with 30 breast scans is the most compact one, followed by the models constructed from 60 and 90 breast scans and the RBSM.
On the other hand, the model built from only 30 breast scans shows the worst generalization ability (about 1.07 mm when using all 29 PCs available).
The model containing 90 breast scans and the RBSM achieved very similar results, both showing a generalization error of about 0.6 mm when considering the first 29 PCs.
As expected, whereas the RBSM is the most specific one, the model learned from only 30 breast scans clearly performs the worst. 

\begin{figure}[t]
\includegraphics{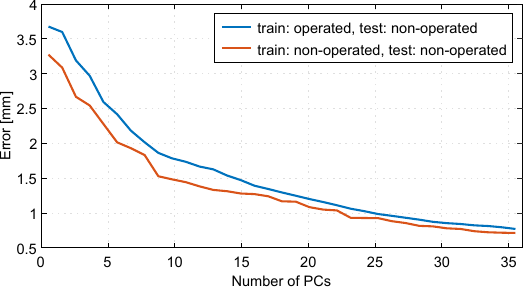}
\caption{Generalization ability of a statistical shape model built from 37 operated breasts, calculated using the 38 non-operated breasts as test set. 
For comparison purposes, a second model was trained with the same number of non-operated breasts. 
Its generalization ability was evaluated on the 38 non-operated breasts using leave-one-out cross-validation.}
\label{fig:exp3_model}
\end{figure}

\subsubsection{Building a statistical shape model from operated breasts}
As already mentioned earlier, a quite big proportion of our 3D breast scans (72 out of 110) result from subjects who already received some kind of breast surgery.
It is thus natural to assume that the RBSM might be biased towards operated breasts and is therefore not able to explain non-operated breasts well.
The goal of this experiment is thus to verify how well a model built from operated breasts is able to reconstruct 3D breast scans from participants that did not undergo breast surgery.
In particular, a statistical shape model is built from a random subset of 37 (out of 72) operated breasts.
Afterwards, generalization ability is evaluated using the 38 non-operated breasts as test set.
Again, to avoid sampling bias, the sampling procedure was repeated three times, and results were averaged.
37 out of the 38 non-operated breasts are the basis for a second model, calculating the generalization ability using leave-one-out cross-validation.

The results are shown in Figure \ref{fig:exp3_model}.
As it can be seen, although our database is obviously biased towards operated breasts, a model built only from operated breasts \textit{is} able to explain non-operated breasts quite well.
Specifically, starting with a generalization error of over 3.5 mm, the generalization ability increases constantly as the number of PCs increase.
Finally, a generalization error of 0.77 mm is achieved when using 36 PCs.
For comparison, the model constructed from 37 non-operated breasts is able to reconstruct non-operated breasts only slightly better with an error of 0.71 mm when using the same number of PCs.

\subsubsection{Generalization ability using clinical parameters}
Our final experiment evaluates the generalization ability of the RBSM using three common, \textit{clinical} parameters typically obtained on the breast.
This way, we aim to show whether or not reconstructions obtained from the RBSM not only well explain unseen objects in general, but also preserve meaningful clinical parameters.
Technically, this is quite similar to the generalization ability as defined by Styner et al. \cite{Sty03}.
However, instead of calculating point-to-point distances between an unseen 3D breast scan and its corresponding reconstruction from the RBSM, we compare three anthropometric measurements obtained on each mesh.
The following three distances between expert-annotated landmarks are measured:
\begin{itemize}
\item sternal notch to left nipple (SN-NL),
\item sternal notch to right nipple (SN-NR), and
\item left nipple to right nipple (NL-NR).
\end{itemize}
Afterwards, the absolute difference between the ground-truth distance taken on the original 3D breast scan and the distance obtained from the reconstruction is calculated for each of the three anthropometric measurements.
Similar to ordinary generalization based on point-to-point distances, we employ leave-one-out cross-validation for all 110 breast scans to calculate our newly defined generalization ability.

\begin{figure}[t]
\includegraphics{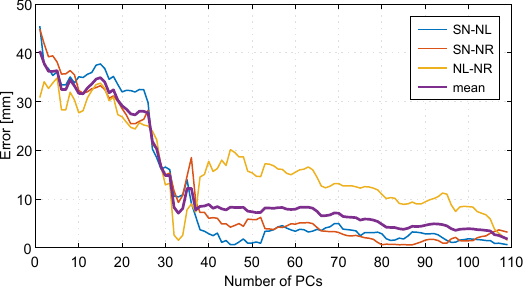}
\caption{Generalization ability of the proposed RBSM, evaluated by comparing anthropometric measurements obtained on an original 3D breast scan and its corresponding reconstruction from the model. 
The average generalization error of the three measurements considered is also shown.}
\label{fig:exp4_model}
\end{figure}

The results are shown in Figure \ref{fig:exp4_model}.
As it can be seen, using less than 30 PCs leads to a high generalization error for all three anthropometric measurements, ranging between 15 mm and 45 mm. 
However, as the number of PCs increases, the generalization error drops significantly and remains low for the SN-NL and SN-NR distances.
Interestingly, only the NL-NR distance increases again up to 20 mm.
In summary, all three anthropometric measurements are best preserved when using 109 PCs, showing an average generalization error of 1.82 mm.
The individual generalization errors when using 109 PCs are 0.66 mm (SN-NL), 3.24 mm (SN-NR), and 1.57 mm (NL-NR).

\section{Applications}
\label{sec:applications}
To further underline the expressiveness of the proposed model, this section demonstrates two exemplary applications for the RBSM that may be used for breast surgery simulation.
The first application (Section \ref{subsec:breast_editing}) is based on the feature editing framework \cite{All03,Bla99} and allows to specifically manipulate clinical parameters on the breast. In the second application (Section \ref{subsec:breast_prediction}) we demonstrate how the RBSM can be used to predict a missing breast from the remaining one by utilizing \textit{posterior} shape models \cite{Alb13}.

\subsection{Breast shape editing}
\label{subsec:breast_editing}
So far, our statistical shape model provides a convenient way to generate new breast shapes by simply varying its parameter $\boldsymbol\alpha\in\mathbb{R}^q$ within a suitable range.
However, a major drawback of this parametrization is the non-interpretability of $\boldsymbol\alpha$ in the sense that it does not correlate with any meaningful features of the breast.
This clearly hinders the generation of breast shapes with certain properties or the ability to alter an existing shape according to some clinical parameters, for instance.
To this end, the feature editing framework proposed by Blanz and Vetter \cite{Bla99} and extended by Allen et al.  \cite{All03} overcomes this drawback by linearly relating features with shape parameters.
Briefly, being $l$ feature values $\{f_{i1},f_{i2},\ldots,f_{il}\}\subset\mathbb{R}^l$ for each individual given and stacked into a feature vector $\mathbf{f}_i\coloneqq(f_{i1},f_{i2},\ldots,f_{il},1)\in\mathbb{R}^{l+1}$, $i=1,2,\ldots,k$, a matrix $\mathbf{F}\coloneqq\left(\mathbf{f}_1,\mathbf{f}_2,\ldots,\mathbf{f}_k\right)\in\mathbb{R}^{(l+1)\times k}$ is defined. 
Arranging the shape parameters of each individual in a matrix $\mathbf{A}\coloneqq\left(\boldsymbol\alpha_1,\boldsymbol\alpha_2,\ldots,\boldsymbol\alpha_k\right)\in\mathbb{R}^{q\times k}$ and following Allen et al. \cite{All03}, $\mathbf{F}$ and $\mathbf{A}$ can be related by means of an unknown transformation matrix $\mathbf{M}\in\mathbb{R}^{q\times(l+1)}$ satisfying
\begin{equation}
\mathbf{MF}\overset{!}{=}\mathbf{A}\quad\Longrightarrow\quad\mathbf{M}=\mathbf{F}^+\mathbf{A},
\end{equation}
where $\mathbf{F}^+$ denotes the Moore–Penrose inverse of $\mathbf{F}$.
Using the matrix $\mathbf{M}$, the shape of an individual can be edited by simply providing new feature values.
If $\Delta\mathbf{f}$ denotes the component-wise difference between a target feature vector and the actual feature vector of an individual, the new shape parameter $\boldsymbol\alpha'$ is obtained as $\boldsymbol\alpha'=\boldsymbol\alpha+\mathbf{M}\Delta\mathbf{f}$.
The edited shape is given by $M(\boldsymbol\alpha')$.

\begin{figure}[t]
\centering
\includegraphics{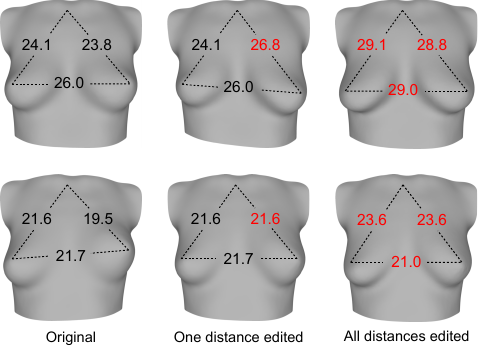}
\caption{Systematic breast shape editing for two examples (top row and bottom row) using the RBSM and three common anthropometric distances as features. 
Note that all measurements are given in centimeters.}
\label{fig:feature_editing}
\end{figure}

For our exemplary application, we use the clinical parameters SN-NL, SN-NR, and NL-NR as features and set $q=k-1$.
Two exemplary results are shown in Figure \ref{fig:feature_editing}.
The RBSM not only allows specific manipulation of important anthropometric distances but also produces at the same time plausible and natural-looking breast shapes, confirmed by the clinical experts involved in this project.

\subsection{Breast shape prediction}
\label{subsec:breast_prediction}
While our first application is designed to alter key clinical parameters of the breast, this section demonstrates how the RBSM can be used to predict a missing breast from the existing one by means of \textit{posterior} shape models \cite{Alb13}.

Given an unseen 3D breast scan in correspondence with the template and represented as $\mathbf{x}\in\mathbb{R}^{3n}$, the area enclosing the missing breast is marked first.
Denote the indices of marked points as $J\subset\mathbb{N}$.
Following Albrecht et al. \cite{Alb13}, the remaining $r\coloneqq n-\left|J\right|$ points provide known observations which we use to compute a posterior mean $\mathbf{\bar{x}}_p\in\mathbb{R}^{3n}$. 
It is the likeliest reconstruction of the missing breast.
Formally, denote as $\mathbf{\bar{x}}_*\in\mathbb{R}^{3r}$ and $\mathbf{U}_*\in\mathbb{R}^{3r\times q}$ the sub-vector and sub-matrix obtained by removing those entries from $\mathbf{\bar{x}}$ and $\mathbf{U}$ corresponding to the selected points in $J$.
Similarly, $\mathbf{x}_*\in\mathbb{R}^{3r}$ represents the target 3D breast scan after removing the marked points.
Then, 
\begin{equation}
\mathbf{\bar{x}}_p=\mathbf{\bar{x}}+\mathbf{U}\left(\mathbf{U}_*^\top\mathbf{U}_*+\sigma^2\mathbf{I}_q\right)^{-1}\mathbf{U}_*^\top\left(\mathbf{x}_*-\mathbf{\bar{x}}_*\right),
\end{equation}
where $\sigma^2\geq 0$ is a small variance accounting for the deviation of $\mathbf{x}_*$ from the model \citep{Alb13} and $\mathbf{I}_q\in\mathbb{R}^{q\times q}$ is the identity matrix.

\begin{figure}[h]
\includegraphics[]{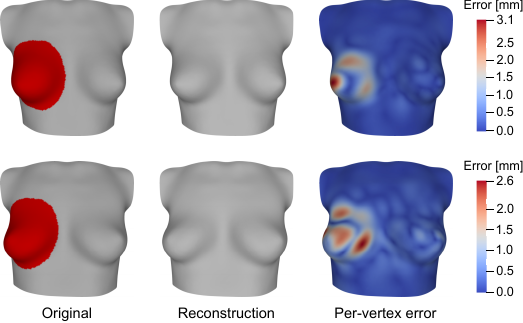}
\caption{Predicting the right breast (marked as \textit{missing} in red) from the left one for two examples (top row and bottom row) using the RBSM.}
\label{fig:breast_prediction}
\end{figure}

Two exemplary results are shown in Figure \ref{fig:breast_prediction}.
Note that, compared to simply mirroring the remaining breast, the prediction obtained from the RBSM is equipped with a statistical probability.
Hence, it may not be the most symmetrical result, but, more importantly, the likeliest and probably most natural one.
This is crucial especially in BRS, where the goal is to reconstruct a breast looking as natural as possible.

\section{Discussion}
\label{sec:discussion}
The following section discusses the most important insights gained from experimental evaluation and also summarizes some limitations of our method.

\subsection{Decoupling between breast and thorax}
The experimental evaluation showed that BPM-based registration is able to decouple the breast region from the thorax well.
The areas with the highest variation in random samples drawn from the RBSM are almost always coincident with the breast region whereas the thorax is kept relatively stiff.
Hence, a variety of different breast shapes can be generated independently from the surrounding areas or without altering the thorax too much, implying a quite well decoupling between breast and thorax.
This can be also verified from Figure \ref{fig:overview_approach}, showing that the unwanted variance of the surrounding regions is properly reduced when using BPM-based registration.
Furthermore, the fact that our feature editing application is able to change only the requested features without altering the shape of the thorax supports our observations.

Finally, we also want to note that there might be an alternative strategy for breaking up the strong correlation between breast and surrounding regions. 
In particular, the framework proposed by Wilms et al. \cite{Wil17} allows constructing statistical shape models with a locality assumption by manipulating the sample covariance matrix.
Distant areas are decoupled by explicitly setting non-zero covariances between those points to zero.
This way, covariances between points in the breast region and points outside could be set to zero, effectively decoupling those regions.
A similar but probably easier way to localize statistical shape models is by using GPMMs as described by Lüthi et al. \cite{Lue18}.
We believe, however, that minimizing variances instead of covariances (between pairs of points) is a more intuitive way to weaken the coupling between breast and thorax. 
Note that there is a strong relation between both approaches since $\left|\cov(x,y)\right|\leq\sqrt{\var(x)}\sqrt{\var(y)}$.

\subsection{Landmarks guiding the registration process}
Using only four landmarks (both nipples as well as both lower breast poles) for non-rigid registration is quite uncommon.
Most of the existing methods need a lot more points to guide the registration process reasonably well.
For instance, the registration of facial 3D scans for the construction of statistical shape models is usually guided by about 60 to 80 landmarks, see e.g. \cite{Boo16,Ger17}.
Due to human anatomy, collecting such a huge amount of landmarks on the female breast is challenging and a fundamental problem of 3D breast scan registration. 
Except for the nipples (and partially the lower breast poles), there are no landmarks that can be \textit{reliably} detected only through visual inspection and consistently across a wide range of differently-shaped breasts.

Besides anatomical points, additional landmarks based on easy-detectable, non-bony structures may be used.
This way, the lower breast pole was simply defined as \textit{the lowest (most caudal) point of the breast}.
From a medical point of view, this is indeed a valid definition of a landmark that is also used in clinical practice.
Based on our collected data, however, it turns out that this definition is not sufficiently precise to be used for registration purposes.
The actually determined lower breast poles were oftentimes \textit{not} accurately located at the lowest point of the breast.
As a result, the initially detected points had to be manually refined in about 40\% of the subjects.

Another fundamental issue is the non-existence of easy-detectable landmarks on the lateral part of the breast.
The absence of those points poses a serious problem in our 3D breast scan registration.
In particular, when not already quite well aligned initially, the template has no chance of being pulled laterally due to the missing guidance.

To conclude, detecting landmarks on 3D breast scans reliably is challenging.
Ultimately, since a physical examination is required to detect anatomical landmarks, automatic landmark detection algorithms based on 3D breast scans (only capturing the surface of the body) are almost impossible to develop.
Additionally, it is important to note that even in some participants clearly detectable landmark points would be present, one has to keep in mind that those landmarks always need to be consistently located across \textit{all} subjects.
By means of the proposed registration pipeline, however, we demonstrated that 3D breast scan registration \textit{is} generally possible by providing only four landmarks.

\subsection{BPM-based registration}
Our quantitative evaluation clearly indicates that BPM-based registration produces superior results than registration without BPMs, affecting both, registration \textit{and} model quality.
When using BPMs, only a little distortion was introduced. 
In addition, although only four landmarks are provided, pretty well correspondences are established. 
We observed only minor correspondence errors when randomly sampling from the RBSM. 

Regarding the EBF-based representation of the BPMs, we can conclude that EBFs are quite well suited to capture the natural teardrop shape of the breast.
However, in some cases, we observed that the current BPMs are not always able to properly capture the \textit{entire} breast region, especially if the breast does not follow a typical teardrop shape.
The reason is the restriction of the covariance matrices (occurring in the Mahalanobis distance) to be diagonal, pulling off a lot of flexibility for the BPM to capture unusual breast shapes.
We expect that, if the full covariance matrix would be specified, better BPMs can be constructed. 
However, this will clearly hinder the automatic parameter selection as 30 values need to be carefully determined for each 3D breast scan.

Lastly, it is important to note that our rigid alignment will completely fail to estimate the correct transformation if the initial alignment is bad.
This is a well-known behavior of the ICP algorithm as it does not optimize for a globally optimal transformation.
Since our 3D breast scans are initially quite well aligned and consistently oriented due to the standardized data acquisition using photogrammetry, we did not run into this problem. 

\subsection{3D breast scan database}
Although our 3D breast scan database contains nearly twice as many operated than non-operated breasts (72 operated, 38 non-operated), the RBSM is likely to show only a minor bias towards operated breasts.
This is supported by the fact that a model built from a randomly chosen subset of 37 operated breasts is able to reconstruct non-operated breasts with an almost similar error than a model trained \textit{and} tested on the same number of non-operated breasts. 
Specifically, an absolute difference of 0.06 mm between both reconstructions was reported.
Intuitively, this suggests that the operated breasts contained in our database look quite natural (or, at least, similar to non-operated ones), making it challenging to decide whether a particular patient received breast surgery or not.
It is, however, important to note that this experiment was conducted by considering only a subset of 37 operated breasts.
This is because otherwise, a model built from an equal amount of non-operated breasts could not be constructed for comparison purposes.
On the other hand, the RBSM contains 72 operated breasts.
However, even if the absolute difference between a reconstruction obtained from a model trained on operated breasts and one that includes only non-operates breasts scales \textit{exponentially} with the number of operated breasts considered, an absolute difference of 0.23 mm would be reported for the RBSM.
It is thus valid to assume that the bias introduced due to operated breasts will not hinder the RBSM from explaining non-operated breasts sufficiently well.

Regarding the amount of data, our experiments confirmed what was expected: the more data, the better. 
The RBSM (containing all 110 breast scans) clearly outperforms the other models trained only on a subset of the data.
Our experiments do not allow a conclusion about whether or not 110 breast scans are already enough in order to represent the vast amount of possible breast shapes well.
However, the difference between 90 and 110 breast scans in terms of generalization and specificity is quite small (see Figure \ref{fig:exp2_model}) strengthening the impression of being near convergence.

Besides, we believe that in order to provide better pose-standardized training data, a more advanced technical set-up is needed in which the movement of arms and shoulders is further constrained. 
This, however, would lengthen the whole data acquisition which might then become unfeasible to be applied in a clinical environment.

\section{Conclusion and outlook}
\label{sec:conclusion}
This paper proposed the RBSM -- the first publicly available 3D statistical shape model for the female breast, learned from 110 breast scans. 
Our extensive evaluation reveals a generalization ability of 0.17 mm and a specificity error of about 2.8 mm when using all 109 PCs available.
Together with the model, a fully automated, pairwise surface registration pipeline was presented which requires only four landmarks to guide the registration process.
In order to break up the strong coupling between breast and thorax and thus being able to capture the actual breast shape as isolated as possible, we proposed to minimize the unwanted variance of the surrounding regions by means of a novel concept called BPMs.
Defined over a surface mesh and based on EBFs, BPMs allow for a probabilistic definition of the breast region and hence overcomes the difficulty in determining an exact delineation of the breast.
Later, BPMs are incorporated into the registration pipeline in order to align the template as accurately as possible inside the breast region and only roughly outside.
With our experiments, we could show that this strategy effectively reduces the unwanted variance outside the breast region and hence, decouples the breast from the surrounding areas very well.
As a result, the RBSM is able to generate a variety of different breast shapes quite independently from the thorax.

In our future work, we ultimately plan to combine physically motivated deformable models and statistical shape models of the breast in order to enable more realistic and statistically plausible surgical outcome simulation for BRS. 
The proposed RBSM is seen as a first step towards this goal.
Moreover, although the results of the proposed sample applications look visually pleasing and promising from a medical point of view, further evaluation is needed in order to assess the practical impact.

Besides the two exemplary applications shown in this paper, there is room for plenty of different applications, extensions, and improvements of our model.
Inspired by recent work of Göpper et al. \cite{Goe20}, an interesting future application could include a thoracic wall into the RBSM to enable breast volume estimation from 3D surface scans.
Furthermore, since our model is generative, it could be conveniently used for data augmentation in the machine learning domain.
The RBSM may also be used as a prior for surface registration algorithms, effectively reducing the search space of possible deformations.
Following the ideas of Booth et al. \cite{Boo16}, so-called \textit{bespoke models} (statistical shape models trained from a dedicated subset of the training set) may be built, for example for different classes of BMIs, ptosis grades, or breast volumes. Additionally, instead of one global model accounting for the whole breast, two individual models for the left and right breast could be built and subsequently combined.
Lastly, our rigid alignment needs to be improved to be able to deal with an arbitrary initial alignment.
For instance, more robust ICP variants such as Go-ICP \cite{Yan16} may be used or advanced feature detectors (based on heat or wave kernel signatures \cite{Aub11,Sun09} or multi-scale mean curvatures \cite{Pan10}, for instance) combined with a robust outlier detection to find reliable correspondences for transformation calculation. 

\section*{Declaration of Competing Interest}
The authors declare that they have no known competing financial interests or personal relationships that could have appeared to influence the work reported in this paper.

\section*{Acknowledgments}
We thank our colleagues from the University Center of Plastic, Aesthetic, Hand and Reconstructive Surgery at the University Hospital Regensburg and the St. Josef Hospital Regensburg for their help during data collection.
This research did not receive any specific grant from funding agencies in the public, commercial, or not-for-profit sectors.

\bibliographystyle{abbrv}
\bibliography{manuscript}

\appendix
\section{Non-linear optimization of $F$ using AM}
\label{app:solve_am}
In this section, we briefly summarize the alternating minimization (AM) approach used to solve the non-linear optimization problem in (\ref{eq:non_rigid_opt_problem}).
Before we proceed by adapting the AM method to our specific problem, we note that the regularization term $F_R$ is actually supposed to have three parameters. 
Besides the new points $\mathbf{P}'$, the scales $\{s_1,s_2,\ldots,s_n\}$ as well as rotations $\{\mathbf{R}_1,\mathbf{R}_2,\ldots,\mathbf{R}_n\}$ are also unknown.
Consequently, the optimization procedure needs to take into account all three variables.
AM does this by independently optimizing for each variable while the remaining variables are fixed.

\subsection{Optimization for scale and rotation}
To begin with, we first perform some algebraic manipulations to simplify further calculations.
Additionally, according to Horn et al. \cite{Hor88}, however, using the formulation of (\ref{eq:asap}) leads to an asymmetry in the determination of the optimal scale factor during the optimization process.
That is, if we transform the surface $\mathcal{S}$ into $\mathcal{S}'$, then the inverse scaling factor from $\mathcal{S}'$ to $\mathcal{S}$ is in general \textit{not} $1/s$ as one would expect.
We thus use the following symmetric version as suggested by Horn et al. \cite{Hor88}, which is given by
\begin{equation}
\frac{1}{\sqrt{s_i}}\left(\mathbf{p}'_j-\mathbf{p}'_k\right)=\sqrt{s_i}\mathbf{R}_i\left(\mathbf{p}_j-\mathbf{p}_k\right)\quad\quad\forall(j,k)\in E_i.
\end{equation}
Using the fact that $\left\Vert\mathbf{Q}-\mathbf{R}\right\Vert^2_F=6-2\tr\left(\mathbf{R}^\top\mathbf{Q}\right)$ for all $\mathbf{Q},\mathbf{R}\in\SO(3)$, we can finally rewrite (\ref{eq:casap_regularization}) into
\begin{multline}
F_R(\mathbf{P}')=\\
\frac{1}{2}\sum_{i=1}^n w_i\left[\sum_{(j,k)\in E_i}w_{jk}\left(\frac{1}{s_i}\left\Vert\mathbf{e}'_{jk}\right\Vert^2_2-2\left(\mathbf{e}'_{jk}\right)^\top\mathbf{R}_i\mathbf{e}_{jk}+s_i\left\Vert\mathbf{e}_{jk}\right\Vert^2_2\right)\right.\\
\left.+3\lambda\sum_{l\in N_i}w_{il}-\lambda\sum_{l\in N_i}w_{il}\tr\left(\mathbf{R}_l^\top\mathbf{R}_i\right)\right],
\end{multline}
where we defined $\mathbf{e}'_{jk}\coloneqq\mathbf{p}'_j-\mathbf{p}'_k$ and $\mathbf{e}_{jk}\coloneqq\mathbf{p}_j-\mathbf{p}_k$.
We will use this symmetric representation of $F_R$ for the calculation of the optimal scale and rotation, both presented next.

\textbf{Scale.}
Taking partial derivative
\begin{equation}
\frac{\partial F_R}{\partial s_i}=\frac{w_i}{2}\sum_{(j,k)\in E_i}w_{jk}\left(-\frac{1}{s_i^2}\left\Vert\mathbf{e}'_{jk}\right\Vert^2_2+\left\Vert\mathbf{e}_{jk}\right\Vert^2_2\right)
\end{equation}
and setting it to zero yields
\begin{equation}
s_i=\sqrt{\frac{\sum\limits_{(j,k)\in E_i}w_{jk}\left\Vert\mathbf{e}'_{jk}\right\Vert^2_2}{\sum\limits_{(j,k)\in E_i}w_{jk}\left\Vert\mathbf{e}_{jk}\right\Vert^2_2}}
\end{equation}
since $w_i,s_i>0$.

\textbf{Rotation.}
Next, we solve for the optimal rotation $\mathbf{R}_i$.
Dropping terms of $F_R$ that do not depend on $\mathbf{R}_i$, using the fact that $\tr\left(\mathbf{vw}^\top\right)=\mathbf{w}^\top\mathbf{v}$ for all $\mathbf{v},\mathbf{w}\in\mathbb{R}^n$, and exploiting some well-known facts of the trace, we are remained with
\begin{equation}
\begin{split}
\mathbf{R}_i&=\argmin_{\mathbf{R}_i\in SO(3)}-\left(\sum_{(j,k)\in E_i}w_{jk}\tr\left(\mathbf{R}_ie_{jk}\left(\mathbf{e}'_{jk}\right)^\top\right)+\frac{\lambda}{2}\sum_{l\in N_i}w_{il}\tr\left(\mathbf{R}_i\mathbf{R}_l^\top\right)\right)\\
&=\argmax_{\mathbf{R}_i\in SO(3)}\tr\Bigg(\mathbf{R}_i\Bigg(\underbrace{\sum_{(j,k)\in E_i}w_{jk}\mathbf{e}_{jk}\left(\mathbf{e}'_{jk}\right)^\top+\frac{\lambda}{2}\sum_{l\in N_i}w_{il}\mathbf{R}_l^\top}_{\eqqcolon\mathbf{S}_i\in\mathbb{R}^{3\times 3}}\Bigg)\Bigg)\\
&=\argmax_{\mathbf{R}_i\in SO(3)}\tr(\mathbf{R}_i\mathbf{S}_i).
\end{split}
\end{equation}
This problem can be efficiently solved using a singular value decomposition of $\mathbf{S}_i$, see e.g. \cite{Sor17}.

\subsection{Optimization for new points}
To solve for the new points $\mathbf{P}'$ we first differentiate $F$ w.r.t. $\mathbf{P}'$ and then set the derivative to zero.
Taking derivative of $F$ leads to
\begin{equation}
\nabla F(\mathbf{P}')=\nabla F_D(\mathbf{P}')+\alpha\nabla F_R(\mathbf{P}')+\beta\nabla F_L(\mathbf{P}').
\label{eq:gradient_points}
\end{equation}
The first and the last term are obvious since they are just ordinary least squares objectives.
The derivative of the regularization term in the middle of (\ref{eq:gradient_points}) is a bit more involved.
We have
\begin{multline}
\frac{\partial F_R}{\partial\mathbf{p}'_i}=\\
2w_i\sum_{j\in N_i}w_{ij}\left(\left(2+\frac{|G|}{2}\right)\mathbf{e}'_{ij}-\left(s_i\mathbf{R}_i+s_j\mathbf{R}_j+\frac{1}{2}\sum_{k\in G}s_k\mathbf{R}_k\right)\mathbf{e}_{ji}\right)\,,
\end{multline}
where we defined $G\coloneqq\{v\in N_i:(j,i)\in E_v\}\subset V$. 
Intuitively, the set $G$ contains all vertices $v\in N_i$ \textit{opposite} to the edge $(j,i)$.
Note that for triangle meshes and non-boundary edges $\left|G\right|=2$, for boundary edges $\left|G\right|=1$.
Setting derivative to zero yields
\begin{equation}
w_i\sum_{j\in N_i}w_{ij}\mathbf{e}'_{ij}=w_i\underbrace{\frac{1}{2+\frac{|G|}{2}}\sum_{j\in N_i}w_{ij}\left(s_i\mathbf{R}_i+s_j\mathbf{R}_j+\frac{1}{2}\sum_{k\in G}s_k\mathbf{R}_k\right)\mathbf{e}_{ji}}_{\eqqcolon\mathbf{h}_i\in\mathbb{R}^3}
\end{equation}
and hence
\begin{equation}
\nabla F_R(\mathbf{P}')=\mathbf{0}\quad\Longleftrightarrow\quad\mathbf{WLP}'=\mathbf{WH},
\end{equation}
where $\mathbf{W}\coloneqq\diag(w_1,w_2,\ldots,w_n)\in\mathbb{R}^{n\times n}$, $\mathbf{H}\coloneqq(\mathbf{h}_1,\mathbf{h}_2,\ldots,\mathbf{h}_n)\in\mathbb{R}^{n\times 3}$, and $\mathbf{L}\in\mathbb{R}^{n\times n}$ is the Laplacian matrix, see e.g. \cite{Bot10}.
Putting everything together finally yields the following sparse linear system
\begin{equation}
\nabla F(\mathbf{P}')=\mathbf{0}\quad\Longleftrightarrow\quad
\underbrace{\begin{pmatrix}
\mathbf{C}\\
\alpha\mathbf{WL}\\
\beta\mathbf{D}
\end{pmatrix}}_{\eqqcolon\mathbf{A}}
\mathbf{P}'=
\underbrace{\begin{pmatrix}
\mathbf{CQ}\\
\alpha\mathbf{WH}\\
\beta\mathbf{Q}_L
\end{pmatrix}}_{\eqqcolon\mathbf{B}}
\end{equation}
with $\mathbf{A}\in\mathbb{R}^{3n\times n}$ and $\mathbf{B}\in\mathbb{R}^{n\times 3}$. This can be quite efficiently solved using iterative algorithms or direct solvers based on Cholesky decomposition, for example.

\section{Parameters for registration}
\label{app:params}
This section lists relevant parameters we used during rigid and non-rigid registration.

\textbf{Rigid alignment.}
The template and target BPMs are thresholded using a value of 0.2. Our modified ICP algorithm terminates if the relative change of MSE between template and target is less than 0.001 or a maximum of 150 iterations is reached.

\textbf{Non-rigid alignment.}
For all three phases, initial, coarse, and fine fitting $\alpha=250$ and $\beta=10^3$ was used. 
However, during initial fitting, no BPMs are used and only one iteration is performed.
For the coarse and fine fitting steps, $\alpha$ is divided by 1.1 and 2 in each iteration, respectively.
As a termination criterion we utilized the distance term $F_D$ and terminate if its value becomes less than 1.9 or if $\alpha<1$. 

\end{document}